\documentclass[sigconf]{aamas}

\usepackage{balance} % for balancing columns on the final page

\usepackage{orcidlink}

\usepackage{xspace}
\usepackage[capitalise, nameinlink]{cleveref}
\newcommand{\acro}{\textsc{Among Us}\xspace} %

\makeatletter
\gdef\@copyrightpermission{
  \begin{minipage}{0.2\columnwidth}
   \href{https://creativecommons.org/licenses/by/4.0/}{\includegraphics[width=0.90\textwidth]{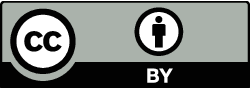}}
  \end{minipage}\hfill
  \begin{minipage}{0.8\columnwidth}
   \href{https://creativecommons.org/licenses/by/4.0/}{This work is licensed under a Creative Commons Attribution International 4.0 License.}
  \end{minipage}
  \vspace{5pt}
}
\makeatother

\setcopyright{ifaamas}
\acmConference[AAMAS '25]{Proc.\@ of the 24th International Conference
on Autonomous Agents and Multiagent Systems (AAMAS 2025)}{May 19 -- 23, 2025}
{Detroit, Michigan, USA}{Y.~Vorobeychik, S.~Das, A.~Nowé  (eds.)}
\copyrightyear{2025}
\acmYear{2025}
\acmDOI{}
\acmPrice{}
\acmISBN{}

\acmSubmissionID{331}

\title[Social Deduction LLMs]{Training Language Models for Social Deduction with Multi-Agent Reinforcement Learning}

\author{Bidipta Sarkar~\orcidlink{0000-0002-0584-3504}}
\affiliation{
  \institution{Stanford University}
  \city{Stanford}
  \country{United States of America}}
\email{bidiptas@cs.stanford.edu}

\author{Warren Xia~\orcidlink{0009-0005-4776-5785}}
\affiliation{
  \institution{Stanford University}
  \city{Stanford}
  \country{United States of America}}
\email{waxia@cs.stanford.edu}

\author{C. Karen Liu~\orcidlink{0000-0001-5926-0905}}
\affiliation{
  \institution{Stanford University}
  \city{Stanford}
  \country{United States of America}}
\email{karenliu@cs.stanford.edu}

\author{Dorsa Sadigh~\orcidlink{0000-0002-7802-9183}}
\affiliation{
  \institution{Stanford University}
  \city{Stanford}
  \country{United States of America}}
\email{dorsa@cs.stanford.edu}

\begin{abstract}
Communicating in natural language is a powerful tool in multi-agent settings, as it enables independent agents to share information in partially observable settings and allows zero-shot coordination with humans. 
However, most prior works are limited as they either rely on training with large amounts of human demonstrations or lack the ability to generate natural and useful communication strategies. 
In this work, we train language models to have productive discussions about their environment in natural language without any human demonstrations. We decompose the communication problem into \textit{listening} and \textit{speaking}. Our key idea is to leverage the agent's goal to \textit{predict useful information about the world} as a dense reward signal that guides communication. Specifically, we improve a model's listening skills by training them to predict information about the environment based on discussions, and we simultaneously improve a model's speaking skills with multi-agent reinforcement learning by rewarding messages based on their influence on other agents. To investigate the role and necessity of communication in complex social settings, we study an embodied social deduction game based on \acro, where the key question to answer is the identity of an adversarial imposter. We analyze emergent behaviors due to our technique, such as accusing suspects and providing evidence, and find that it enables strong discussions, doubling the win rates compared to standard RL. We release our code and models at \url{https://socialdeductionllm.github.io/}.
\end{abstract}



\newcommand{\BibTeX}{\rm B\kern-.05em{\sc i\kern-.025em b}\kern-.08em\TeX}

\begin{document}

%%% The following commands remove the headers in your paper. For final 
%%% papers, these will be inserted during the pagination process.

\pagestyle{fancy}
\fancyhead{}

%%% The next command prints the information defined in the preamble.

\maketitle 

%%%%%%%%%%%%%%%%%%%%%%%%%%%%%%%%%%%%%%%%%%%%%%%%%%%%%%%%%%%%%%%%%%%%%%%%

\begin{figure*}[t]
  \centering
  \includegraphics[width=\textwidth]{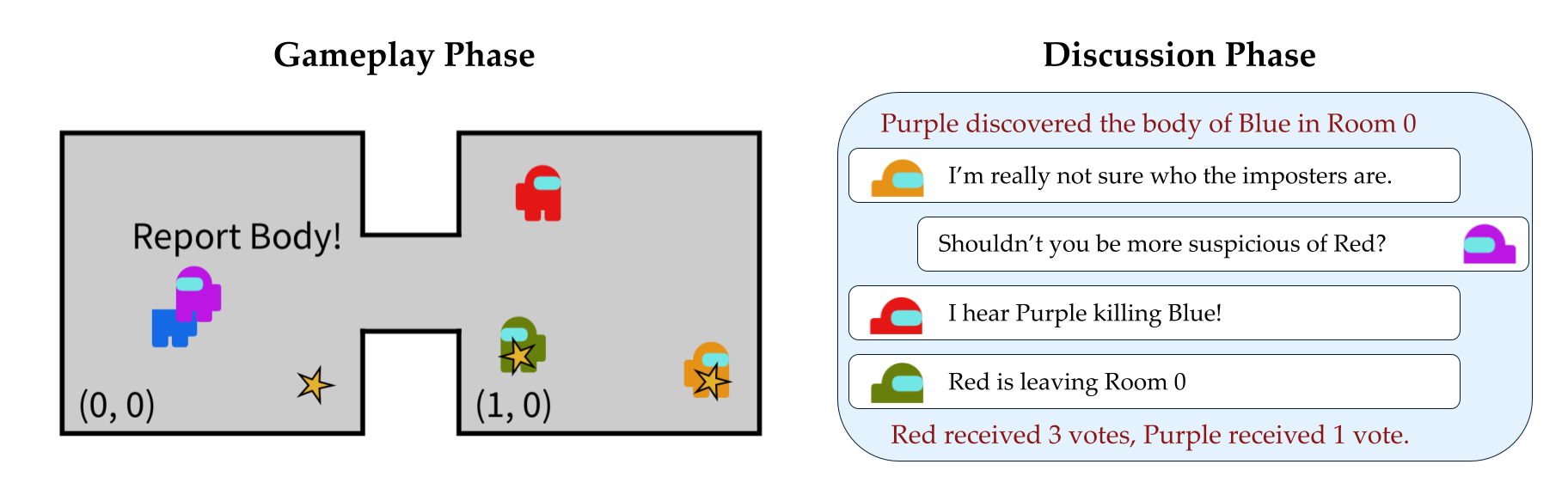}
  \caption{Examples of the gameplay and discussion phases of \acro. During gameplay, all agents navigate a 2D grid environment (a 1-by-2 grid in this case, with two rooms at (0,0) and (1,0)), where agents can see everything in their same room. Here, the red, green, and yellow agents are in room (1,0), and the purple and blue agents are in room (0,0). Crewmates can perform tasks (indicated by the stars -- in this example there are 3 tasks), while imposters kill crewmates. Here, the orange and green agents are working on tasks. Agents can also report dead bodies, as the purple agent is currently doing, which initiates the discussion phase. During discussion phases, agents leverage large language models to generate free-form messages guided by our framework encouraging effective speaking and listening within the crewmates and finally vote out a suspected imposter. The example discussion shown on the right is based on a generated discussion from our trained models.}
    \label{fig:amongus_diagram}
  \Description{Overview of Among Us Components}
\end{figure*}

\section{Introduction}

% High level problem and challenges: We want AI in multi-agent settings to learn to communicate. Helpful for human-AI coordination and partial observability. However, it's not easy to know "what" information is relevant; it may depend on the task.

A longstanding goal of multi-agent artificial intelligence is the development of independent agents that can communicate using a shared language. Communication is especially necessary in ``partially observable'' settings, where each agent only has a limited view of the world and therefore benefits from sharing knowledge with other agents to achieve its goal. In particular, ``social deduction'' games are settings where each agent's goal is to deduce information about the environment by communicating with other agents -- requiring each player to learn how to parse messages from other players while effectively sharing important information needed for game completion.
% For instance, eusocial colonies of animals communicate to reveal the locations of food sources. We can model these settings as ``social deduction'' games, where agents have a goal to deduce information about their environment by communicating with one another. These settings are challenging because each player needs to parse messages from other players along with knowing how to effectively share important information.
% Social deduction games are important settings, because success in these games is strongly correlated with how well agents can understand their environment and communicate useful information to teammates. 

In this work, we study the hidden-role game of \acro~\citep{amongus} as a specific instance of a challenging social deduction game to investigate the importance of communication, illustrated in \cref{fig:amongus_diagram}. Hidden-role games~\citep{carminati2023hiddenrole, kopparapu2022hidden} are a class of environments where players are split into an uninformed majority and a smaller informed hidden subteam, which we refer to as \textit{crewmates} and \textit{imposters} respectively. These two teams are adversaries, resulting in a zero-sum game, where the goal of the crewmates is to deduce the identity of imposters to vote them out. Unlike other popular hidden role games such as the game of Mafia~\citep{Braverman_2008}, where statements from players are unfalsifiable,
% Typically, imposters know the team assignments of all players, while crewmates need to deduce the roles of other players through discussions, i.e., using free form natural language. These games fundamentally challenge players to be mindful of other agents, determine who is suspicious, and effectively communicate their knowledge to other players -- as their reward from winning the game is directly affected by the actions and beliefs of other players. 
\acro is based in a 2D embodied environment, allowing discussions and intuitions to be grounded in specific observations. In the game, crewmates try to complete an assigned set of tasks scattered across the environment while imposters try to kill all crewmates. If a player reports the corpse of an eliminated crewmate -- killed by an imposter -- the game moves to a discussion phase with a free-form chat followed by a voting period, where all players vote to eject a suspected imposter. For crewmates, success in the discussion phase would mean correctly voting out the imposter, while success for imposters means avoiding suspicion from the crewmates to continue staying in the game as long as possible. This highlights the importance of communication during the discussion phase as crewmates need to learn to effectively utilize the discussion phase to vote out imposters in an adversarial setting.
For the rest of this paper, we study the game of \acro from the perspective of crewmates attempting to perform tasks, identify imposters, and win the game.
% Crewmates must learn to effectively utilize the discussion phase to vote out imposters to win the game, so we are interested in improving the discussion abilities of crewmates in adversarial settings.
% Relevant related work: MARL is effective when just learning actions. However, prior work assumed access to large dataset of human demonstrations to learn communcation.

In multi-agent environments, an effective technique for training strong cooperative and competitive agents is multi-agent reinforcement learning (MARL), which enables artificial agents to achieve superhuman levels of performance in competitive games such as StarCraft~\citep{Vinyals2019GrandmasterLI}, and cooperative games such as Overcooked~\citep{carroll2019utility, sarkar2023diverse} and Hanabi~\citep{hu2020other}. However, in settings where communication in natural language is necessary, existing MARL techniques often struggle as they require large datasets of task-specific human communication data to perform on-par with humans~\citep{doi:10.1126/science.ade9097}. This fundamentally limits the agents' ability to communicate at human-level and is not practical for learning in settings where these datasets do not readily exist. The game of \acro falls into this category, where communication is necessary to reason and progress in the game. Therefore, we would like to find an approach that learns to communicate effectively and convincingly without requiring large amounts of task-specific human data. 
% Key insight: Our rewards for the quality of discussion can be grounded directly based on the prediction of scene information, reinforcing the signal for discussion quality
However, the major challenge in learning to communicate without access to large amounts of human data is that novice agents do not have a strong signal for understanding the helpfulness of the messages they send (\emph{speaking}) or for learning the meaning of messages from other players (\emph{listening}). In particular, the sparse reward signal the agents receive when winning the game is not informative enough to reinforce high-quality discussions between agents. Our key insight is that we can leverage the agents' instrumental goal of \emph{predicting useful information about the world} -- e.g., the identity of imposters -- as a dense reward to provide a higher-quality signal that can enable more informative communication during the discussion phase and potentially higher performance policies.

We propose an approach that rewards a message generated during the discussion phase based on how the other crewmates' beliefs on the identity of the imposter changes. Each crewmate wants to send messages that help other crewmates be more certain about the true identity of the imposter. However, this only explains how to learn to ``speak'' assuming that the other agents can appropriately update their belief about the world given a message. We also need to ensure the agents know how to ``listen'' and update beliefs appropriately. To encourage this, we additionally add an \emph{imposter prediction} signal to guide the agent's learning to predict the true identity of the imposter after each message. By training agents to speak and listen effectively, we enable the agents to self-improve their discussion abilities.
Further, to encourage \emph{listening} and \emph{speaking} in natural language during the discussion phase of the game, we tap into the power of large language models (LLMs), unspecialized models trained with large amounts of human language data. Specifically, we initialize crewmates as LLMs capable of communicating via natural language.  
Recent advances in foundation models have demonstrated some reasoning abilities~\citep{bubeck2023sparks, generalpatternmachines2023}, including understanding social scenarios~\citep{kwon2023toward}, but even the strongest language models today are weak at self-critiquing~\citep{stechly2023gpt4} or performing theory of mind reasoning~\citep{shapira2023clever}, limiting their ability to improve their listening skills based on their own feedback. However, by training LLMs within our proposed framework of encouraging listening and speaking with auxiliary dense rewards for helping other crewmates vote out the correct imposter, we overcome this limitation, enabling the self-improvement of these models over time.

% To evaluate our framework, we analyze the success rate of crewmates against pretrained imposters and validate their robustness across multiple configurations of \acro, varying the size of the environment, number of tasks per crewmate, and number of players, and find strong generalization in novel test-time configurations. 
To evaluate our framework, we analyze the success rate of crewmates against both pretrained and adaptive imposters, and find that crewmates form a robust communication strategy. We find that our technique results in emergent behavior commonly found in real games of \acro between humans, such as directly accusing players and providing evidence to help other crewmates~\cite{lai-etal-2023-werewolf}. We also find that our augmentation to discussions results in two times higher success rates relative to standard RL along with over three times higher success rates relative to base models that are over four times larger than our models, highlighting the importance of our discussion strategies.

\section{Related Work}

In this section, we review related work on emergent communication, prior works that use language models as agents in embodied settings, and past works integrating language models with RL.

\smallskip

% By pretraining on internet-scale data, large language models (LLMs) contain a vast amount of semantic knowledge, enabling them to be effective agents across many domains. In this section, we reflect on prior work that uses language models for sequential decision making tasks. 
\noindent \textbf{Emergent Communication.} A major topic in MARL is emergent communication between agents, especially in the context of reference games and repeated reference games, where a speaker knows the ground-truth answer to a question (e.g., a specific image out of a set of images that needs to be referred to). Then, the speaker needs to communicate to the listener, who later needs to choose the item being referenced either over one or repeated interactions.
Prior work has shown that humans tend to quickly adapt to such tasks~\cite{mccarthy2021learning}, naturally using theory of mind reasoning to determine the intents of speakers~\cite{FRANK201480}. Further, \citet{hawkins-etal-2020-continual} showed that language models can also learn to adapt to human conventions via continual learning. Without using human natural language data, \citet{Lazaridou2016} and \citet{Havrylov} use symbolic cheap-talk signals to solve referential games. Our framework of social deduction games, however, is more challenging as each agent does not know the ground truth answer, so teams must communicate to collectively learn the answer. Therefore, our domain does not have as clear of a distinction between ``speakers'' who have knowledge and ``listeners'' who need to gain answers as agents in social deduction games must play both roles.

\smallskip

\noindent \textbf{Language Models Agents.} A large body of prior work use LLMs' access to internet scale data for task planning and decision making. In robotics, prior works explore how language models can be used to plan out a sequence of high-level primitives given an instruction in natural language~\cite{saycan2022arxiv, huang22a, Lin2023}. %\citet{gdm2024autort} also uses foundation models to control a fleet of robots for collecting diverse data. 
In a virtual gaming setting, \citet{Park2023GenerativeAgents} uses ChatGPT to simulate members of a small virtual town. Although there is no specific task or mechanism for ``training'' these agents, they demonstrate the use of a long-term memory stream to store memories beyond the context length of the language models, enabling the formation of social networks. This technique of having external memory has later been used to learn ``skills'' in a single-player environment~\citep{wang2023voyager} and for coordination in multi-agent environments~\citep{gong2023mindagent}.  These works demonstrate that language models are capable of controlling agents in a wide range of settings, which is key to our motivation to directly use language models as a strong starting point for agents operating in more challenging environments such as social deduction games.

\smallskip

\noindent \textbf{Reinforcement Learning with Foundation Models.}
Some works also combine language models with reinforcement learning. Cicero \citep{doi:10.1126/science.ade9097} is an AI for the game of Diplomacy that uses a dialogue-conditional action model from human actions and trains a dialogue-free model using RL to choose actions. Cicero uses an ``intent'' embedding to connect the dialogue generation and strategic reasoning components. This allows Cicero to communicate with other agents in a way that feels natural to other players, but it prevents the RL model from directly controlling the generated messages, potentially limiting improvements in message quality. Another drawback is that this technique requires a large number of human demonstrations, which may be impractical in many settings.

Foundation models have been effective in both providing rewards and as a base model for policies. \citet{hu2023language} and \citet{kwon2023reward} use language models as reward signals to train a separate network to follow a specific coordination strategy. We similarly use the LLM to provide denser rewards during the discussion phase, but we train the LLM itself instead of a separate policy. 
%\citet{szot2023large} uses a frozen pretrained vision encoder and a language model to train an action decoder with RL to perform robotics tasks based on language instructions. In contrast, we represent both actions and observations in the space of tokens, so we fine-tune the language model directly instead of having a separate action decoder.

Outside of the embodied setting, reinforcement learning has also been key to improving the chat capabilities of LLMs. \citet{ouyang2022training} demonstrates the effectiveness of reinforcement learning from human feedback (RLHF), where a reward model is trained using human feedback and an LLM is fine-tuned using a modification of the PPO algorithm to improve its performance. \citet{yuan2024selfrewarding} extends this by allowing the LLM to be its own reward model and generate its own data for self-improvement, similar to how we use the LLM's own change in beliefs as a reward signal. However, a crucial difference is that our reward model remains grounded in an environment by design due to the imposter prediction training signal. This means that we do not need to rely on the ability of pretrained LLMs to critique their own generations, enabling us to use smaller language models and correct logical errors over time.

\section{Preliminaries}

We model social deduction games, such as \acro, as a variant of the partially observable Markov game (POMG)~\citep{liu2022sampleefficient} that includes a question whose answer must be deduced through interacting with players and the rest of the environment. Our modified POMG can be described by a tuple $(n, \mathcal{S}, \mathcal{A}, \mathcal{P}, r, \mathcal{O}, \gamma, \mathcal{Q}, q)$, where $n$ is the number of players, $\mathcal{S}$ is the joint (hidden) state space and $\mathcal{A}$ is the joint action space. The transition function $\mathcal{P}: \mathcal{S} \times \mathcal{A} \times \mathcal{S} \rightarrow [0, 1]$, is the probability of reaching a state given the current state and joint action. The reward function $r: \mathcal{S} \rightarrow \mathbb{R}^n$, gives a real value reward for each state transition to each player, and $\gamma$ is the reward discount. The observation function, $\mathcal{O}: \mathcal{S} \rightarrow O^n$, generates the player-specific observations from the state.

Our POMG has additional terms for the task of social deduction, which are the set of all possible answers to the deduction problem $\mathcal{Q} \subseteq \mathcal{A}$ and the correct answer $q \in \mathcal{Q}$. In social deduction games, agents will be given opportunities to answer the question as a literal action (i.e. voting in \acro or choosing the correct object in reference games), and at those steps the correct action to take is $q$.

% We model the multi-agent game of \acro as a partially observable Markov game (POMG)~\citep{liu2022sampleefficient}. The POMG can be described by a tuple $(n, \mathcal{S}, \mathcal{A}, \mathcal{P}, r, \mathcal{O}, \gamma)$, where $n$ is the number of players, $\mathcal{S}$ is the joint (hidden) state space and $\mathcal{A}$ is the joint action space. The transition function $\mathcal{P}: \mathcal{S} \times \mathcal{A} \times \mathcal{S} \rightarrow [0, 1]$, is the probability of reaching a state given the current state and joint action. The reward function $r: \mathcal{S} \rightarrow \mathbb{R}^n$, gives a real value reward for each state transition to each player. The observation function, $\mathcal{O}: \mathcal{S} \rightarrow O^n$, generates the player-specific observations from the state. Finally, $\gamma$ is the reward discount.

The trajectory up to time $t$ is defined as a sequence of joint observations and actions: $\tau_t = (o_0, a_0, \dots, a_{t-1}, o_t)$. An individual player only experiences their own action-observation history (AOH), which is defined for player $i$ as $\tau^i_t = (o_0^i, a_0^i, \dots, a_{t-1}^i, o_t^i)$, and they follow a stochastic policy $\pi^i(a^i | \tau^i)$. In the game of \acro, the AOH consists of past observations supplied by the environment, past embodied actions, and all prior discussions with the other players.

% Each player follows a stochastic policy $\pi^i(a^i | \tau^i)$, which is the probability that agent $i$ chooses action $a^i$ given its AOH $\tau^i$. At time $t$, the environment is at state $s_t \in \mathcal{S}$, so the agents receive observations $O(s_t)$ and sample their actions as $a^i_t \sim \pi^i(a^i_t | \tau^i_t)$. The environment generates the next timestep as $s_{t+1} \sim \mathcal{P}(s_t, a_t, s_{t+1})$ and the reward as $r(s_{t+1})$. The discounted reward for player $i$ in a trajectory is $R_i(\tau_T) = \sum_{t=1}^T \gamma^t r_i(s_t)$. The expected reward for a set of policies for the $n$ agents, $\Pi = (\pi^1, \pi^2, \dots, \pi^n)$, is $\mathcal{J}(\Pi) = \mathbb{E}_{\tau \sim (\Pi)}[R(\tau)]$.

\smallskip

\noindent \textbf{Language Models.}
Language models are trained to model the probability of a sequence of discrete tokens, where each token represents a string of natural language. For a sequence of tokens, $W = \{w_0, w_1, \dots, w_k\}$, the probability of the sequence being generated is $p(W) = \prod_{j=0}^k p(w_j | w_{<j})$, so causal language models predict the distribution of the next token conditioned on all prior tokens. 

Our \acro environment is designed such that each observation at time step $t$ is a sequence of tokens $o_t = W^t = \{ w_0, w_1, \dots, w_k \}$ and each action at time step $t$ is a single token $a_t = w_t$, allowing us to use language models as the policy for each agent. The AOH is a sequence of tokens, so language models can sample the next action by predicting the next token in the sequence, constrained to the set of legal actions at that timestep.

In this work, we use the RWKV language model~\citep{peng2023rwkv}, a recurrent language model based on a linear attention mechanism, as the pretrained foundation model. We choose RWKV over more common transformer-based models~\cite{vaswani2023attentionneed}, because the recurrent formulation allows us to generate RL trajectories with a constant time and space complexity per token, and RWKV enables unbounded-context training using truncated backpropagation through time. This is especially important since \acro trajectories often reach tens of thousands of tokens in length per player, which would require significantly more compute for classic attention-based models. Empirically, RWKV has also performed on-par with transformer-based models, especially in decision-making tasks~\cite{dong2024optimizingroboticmanipulationdecisionrwkv} and long-context understanding~\cite{huang2024longsequencemodelmodel}, making it the ideal choice for this study.

% \karen{Need to be explicit about how language models are related to the POMG we described above. Do we use language models to represent agent policies? Do we use large models to the world?}To maintain compatibility with language models, all observations and actions are sequences of \textit{tokens}, which are discrete values interpreted by language models as pieces of words. Not all tokens are always considered valid actions, so the environment provides a list of valid actions at each timestep.

% In this work, we use the RWKV language model~\citep{peng2023rwkv} as the pretrained foundation model. We choose this architecture because it has a constant time complexity to generate a next token regardless of the number of prior tokens, unlike the common transformer-based architectures~\citep{transformers}, where generation time grows linearly with the number of prior tokens. This is preferable for reinforcement learning, since it reduces the time needed to generate trajectories, especially in our \acro setting where full game trajectories can reach tens of thousands of tokens in length.

\section{The Game of \acro}

In this section, we describe the key design decisions of our implementation of the hidden-role game of \acro. Our goal is to create an environment where agents can ground their discussion based on evidence in the environment. A more complete description of the game is in Appendix A.%\cref{app:env}.

% Since the goal of our work is to determine the benefit of cooperative discussions, we measure the success of crewmates as a proxy for our algorithm's ability to learn meaningful discussions in the presence of adversaries such as imposters. 

\smallskip

\noindent \textbf{Role Assignment.} At the start of the game, each player is either assigned as an \textit{imposter} or a \textit{crewmate}. The crewmates are not informed of the identities of the other players, but all imposters are informed of the identities of the other players at the start.

% In our experiments, we mainly consider 5-player settings, where one player is the imposter and the other four players are crewmates.

In our setting, we assign one player to be the imposter and the other $n-1$ players as crewmates. The crewmates are assigned a set of $N$ tasks, scattered across the environment. As an example, $N=3$ in the example in ~\cref{fig:amongus_diagram}.

\begin{figure}
    % \begin{minipage}[b]{.48\textwidth}
         \centering
  \includegraphics[width=\columnwidth]{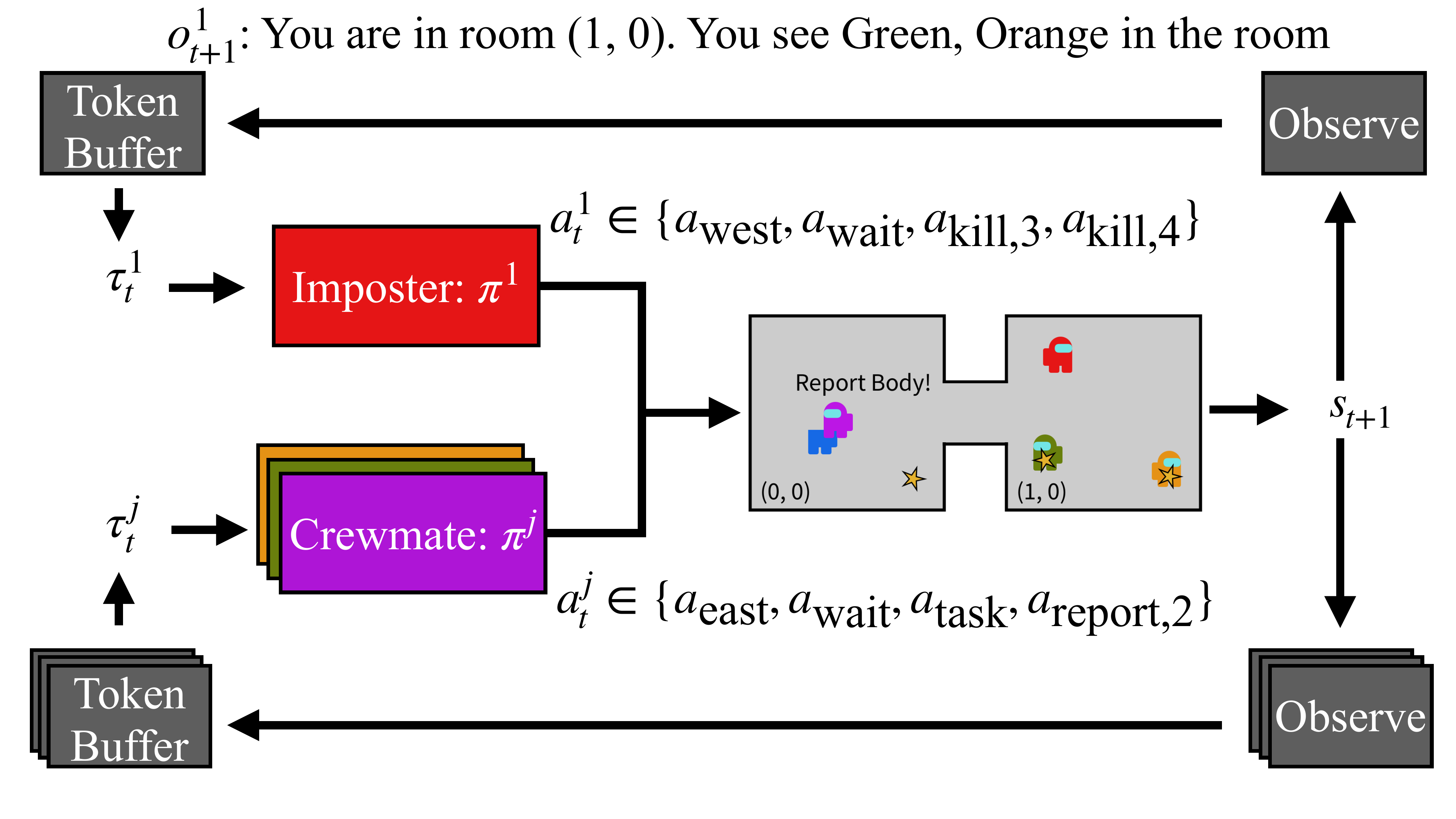}
  \caption{Diagram of the embodied gameplay loop. The environment starts by sending observations to all agents simultaneously and collects tokenized actions from a set of valid actions at each timestep.}
    \label{fig:gameplay}
  %   \end{minipage}%
  %   \hfill
  %   \begin{minipage}[b]{.48\textwidth}
  %       \centering
  %       \includegraphics[width=\textwidth]{figures/discussion_gameplay.pdf}
  % \caption{Diagram of the discussion loop. At each iteration, a selected speaker broadcasts a message to all players, and crewmates report their updated belief on the identity of the imposter to the discussion moderator.}
  %   \label{fig:discussion}
  %   \end{minipage}
    \Description{Embodied Gameplay Loop}
\end{figure}

\smallskip

\noindent \textbf{Gameplay Phase.} During the gameplay phase, players simultaneously move in an embodied environment, receiving observations from the environment and taking actions, as illustrated in~\cref{fig:gameplay}. Players freely move around a $W \times H$ grid of rooms during the gameplay phase, receiving new observations $o_t$ at each time step. All agents can move between adjacent rooms by choosing $a_\text{go to x}$, where x is a cardinal direction, or they can simply wait in the room by choosing $a_\text{wait}$. Crewmates can complete tasks in their current room by choosing $a_\text{task}$, but they are unable to observe the environment for $N_\text{task\_time}$ time steps, i.e., they will not be able to observe if a crewmate is being killed by an imposter while performing a task. Note that tasks are indistinguishable from one another, so we do not have different actions for different tasks. Imposters can kill crewmates by choosing $a_{\text{kill}, j}$ where $j$ is a crewmate in the same room as them, but they have to wait $N_\text{cooldown}$ time steps between killing crewmates. Finally, crewmates can report dead bodies in their room by choosing $a_{\text{report}, j}$, where $j$ is the corpse of player $j$, which initiates the discussion phase.

The set of all valid actions are $a_\text{go to x}$, $a_\text{task}$, $a_{\text{kill}, j}$, $a_{\text{report}, j}$, and $a_\text{wait}$, where x is a cardinal direction and $j$ is the name of a crewmate. The environment provides each player with the subset of actions that are valid at each timestep.

\smallskip

\noindent \textbf{Discussion Phase.} During the discussion phase, we cycle over each player twice in a random order and allow them to say a sentence, $a_\text{talk}$, in natural language. After this discussion, a voting phase begins, where each player votes for one player, $k$, they want to eject by choosing action $a_{\text{vote}, k}$. The player who gets the plurality of votes is ejected. If the imposter is not ejected, the game continues to the next gameplay phase, where crewmates can continue finishing tasks.

Before the discussion starts and between each discussion message, the environment also surveys each crewmate by asking who they would vote for, i.e., by querying them to pick an $a_{\text{vote}, k}$ as if they had to vote immediately. This action has no impact on the POMG, but it will be relevant for our training algorithm.

Note that the set of all voting actions is equal to the set of all possible ``answers'' in the social deduction game ($\mathcal{Q}$), and voting out the correct imposter corresponds to $q$, the correct answer to the social deduction question.

\smallskip

\noindent \textbf{Reward Structure.} \acro is fundamentally a team zero-sum game, so reward is based on whether crewmates or imposters win. If all tasks are completed or the imposter is ejected, the crewmates win with a reward of +1. However, if the number of imposters is ever greater than or equal to the number of crewmates, the imposters win, resulting in a crewmate reward of -1.

% \section{Training LLM Agents to Discuss}
\section{Training LLM Crewmates in \acro}

By defining an environment that only interfaces with players through natural language, we can directly use a language model as the policy $\pi^i(a^i|\tau^i)$ of an agent $i$. The action-observation histories of our agents $\tau^i$ are just strings of natural language, and new observations and actions can simply be appended to the end of the strings. Furthermore, when taking actions $a^i$, the outputs of the language model can be constrained to be one of the legal actions provided by the environment at each timestep. Following this procedure, we construct an agent using a pretrained RWKV language model, which we define as policy $\pi_\text{RWKV}$.

Although the environment is designed to interface nicely with language models, we find that $\pi_\text{RWKV}$ struggles to reason as crewmates in a zero-shot fashion in \acro, with models frequently voting to eject the wrong players. In this section, we describe our procedure for improving the performance of crewmates by enabling them to self-critique and use these scores to improve dialogue generation.

The first two subsections describe how to improve the performance of an individual learning crewmate, first describing a reinforcement learning procedure and then describing how to enhance communication by learning to listen and speak. The third subsection describes how to train the team of crewmates to be robust to adaptive imposters and different policies within the crewmate population. 

%In \cref{world} and \cref{rl}, we apply general techniques for adapting an LLM policy to an environment: self-supervised learning for domain adaptation and reinforcement learning to improve actions. In \cref{listening} and \cref{speaking}, we detail our technique for specifically improving the quality of communication.

\subsection{Reinforcement Learning in \acro}

To train a language model to take more effective actions without expert demonstrations, we can turn to reinforcement learning. Since \acro already provides rewards for winning, we can directly optimize this to produce a model $\pi_\text{RL}$ that minimizes the following loss:
\begin{equation}
    L_\text{RL}(\pi) = -\underset{\tau^i \sim \Pi}{\mathbb{E}}\sum_t \left[\gamma^t r^i_t + \lambda_\text{NL} \log(\frac{\pi(a^i_t | \tau^i_t)}{\pi_\text{RWKV}(a^i_t | \tau^i_t)})\right],
\end{equation}
where $\Pi$ represents the joint policy that has $\pi$ controlling agent $i$, and $\lambda_\text{NL}$ is a hyperparameter controlling the strength of a soft KL constraint regularizing trained models to the base LLM to prevent discussions from moving out of natural language~\cite{ouyang2022training}. Note that the only reward signal is the sparse reward received at the end of the game along with additional rewards for completing tasks. In particular, there is very little signal for the effectiveness of its messages during discussions, which makes utilizing communication very difficult with just RL in practice. This sparse signal also makes identifying the imposter difficult in the multi-agent setting, because voting correctly may still result in a loss and voting incorrectly could result in a win if a plurality of agents vote for the imposter.

\subsection{Enhancing Communication of Crewmates}

To improve beyond the RL baseline, we can take advantage of the social deduction component of the game. In particular, each agent's belief in choosing the correct answer $q \in \mathcal{Q}$ will provide a stronger signal for learning the core components of the game and the means of communication relative to the RL baseline.

In this subsection, we discuss the key contributions of this work. Specifically, we highlight new loss terms to enhance both listening and speaking abilities, enabling crewmates to better utilize discussions.

\smallskip

\noindent \textbf{Listening: Imposter Prediction.}
\label{listening}
Suppose an agent is learning in a environment where it is partnered with expert crewmates who already know how to discuss the game. How can this agent learn to understand the meanings of environment observations and messages from other crewmates?

To effectively discuss the game of \acro, crewmates need to understand who the imposter is given their past observations and the past messages. This prediction task can act as an auxiliary task that can guide the discussion phase to be more grounded and meaningful.
% We frame the imposter prediction problem as a supervised learning problem because the identity of the imposter is known by the end of the game.

We directly train crewmates to improve their reasoning over imposters using the environment's ground truth answer for the identity of the imposter. Specifically, we use the timesteps when the environment directly surveys the players for their beliefs over the imposters, which occurs between discussion messages, as the training signal. Note that this training signal does not specifically require human demonstration data; agents can learn to understand observations and messages from other players using any rollout buffer.

% We desire for the crewmates' beliefs over imposters to be accurate with calibrated confidences. If crewmates are overconfident in their predictions, they may incorrectly vote out the wrong imposter and lose the game. To ensure calibrated confidence throughout the training process of predicting imposters, 
% To address this issue, 
% we use the focal loss, which is shown to reduce sensitivity to misclassified examples
% to learn models that stay calibrated throughout the training process
% ~\citep{mukhoti2020calibrating}.
For every living crewmate, if they are asked to provide their beliefs regarding the identity of the imposter at timestep $t$, the \textit{listening loss} for that timestep is
\begin{equation}
    L_\text{L}(\pi, \tau^i_t) = -\log \pi(q | \tau^i_t),
    \label{eq:focal}
\end{equation}
where $q=a_{\text{vote},j}$ is the action representing choosing the correct imposter $j$, and $\tau^i_t$ is the AOH until timestep $t$, which may include prior discussions.

At the very start of the discussion phase, agents need to reflect on the probabilities of other agents being the imposter based on their observations during the gameplay phase. For instance, if a crewmate directly witnesses a murder, they should be very certain that the murderer is the imposter; our listening loss uses this signal to increase their certainty over the imposter.

% We use this supervised training signal throughout training to ensure that agents continue to correct their beliefs over imposters as their behavior changes. For instance, if an imposter learns to speak deceptively in discussions, this training signal ensures that crewmates learn that imposters can behave in this manner and correct their beliefs accordingly.

% During the voting actions that have an impact on the gameplay, we remove the supervised training signal because we want to allow agents to also choose to ``skip'' if they are uncertain about the identity of the imposter.
By framing the task of identifying imposters using messages and observations as a supervised learning problem, agents learn to understand the meaning of messages, enabling them to vote out the correct imposter. Using this loss term, we can define two new policies. We can directly incorporate the listening loss into the RL policy, giving us the policy $\pi_\text{RL+L}$ that optimizes
\begin{equation}
    L_\text{RL+L}(\pi) = L_\text{RL}(\pi) + \underset{\tau^i \sim \Pi}{\mathbb{E}}\sum_t \lambda_\text{L} L_\text{L}(\pi, \tau^i_t),
\end{equation}
where $\lambda_\text{L}$ is a hyperparameter controlling the strength of the listening loss and is only nonzero on timesteps when the crewmates are asked to predict the identity of the imposter. This enables the model to optimize actions while improving its ability to identify imposters.

We can also define a purely \textit{listening} policy, $\pi_\text{L}$ that incorporates the listening loss without an RL component, therefore optimizing
\begin{equation}
    L_\text{L only}(\pi) = \underset{\tau^i \sim \Pi_\text{rand}}{\mathbb{E}}\sum_t \lambda_\text{L} L_\text{L}(\pi, \tau^i_t),
    \label{eq:base}
\end{equation}
where $\Pi_\text{rand}$ is a joint policy that uses $\pi_\text{RWKV}$ for discussions and chooses gameplay actions uniformly at random. 

\smallskip

\noindent \textbf{Speaking: Reinforced Discussion Learning.}
\label{speaking}
So far, we have developed a policy that can learn to take effective actions in the environment with RL, and can update beliefs based on discussion messages. Now suppose that an agent is partnered with expert crewmates who already know how to parse messages from other players. How can this agent learn to construct helpful messages when it is their turn to speak?

Although our use of a supervised imposter prediction loss allows agents to learn how to interpret messages from other agents in the previous subsection, we cannot directly apply the same idea to learning how to speak as there is no ground truth notion of effective messages. We instead improve the agents' discussion abilities using reinforcement learning. Specifically, we grant rewards to the speaking agent based on the change in living crewmates' beliefs on the true imposter after each message. Formally, let $B_t$ be the sum of all living crewmates' beliefs,
\begin{equation}
    B_t = \sum_{k \in C_t} \pi^k(q | \tau^k_t),
\end{equation}
where the $q$ represents voting out the correct imposter, and $C_t$ is the set of all living crewmates at time $t$. If $t'$ is the previous belief-querying timestep, then the reward for crewmate $i$, who just finished speaking, is $r^s_t$:
\begin{equation}
    r^s_t = B_t - B_{t'}.
\end{equation}
% For imposters, we provide the reward of $-r^c_t$ to incentivize reducing suspicion on themselves.

Intuitively, this reward models the causal effect of each message on the task of predicting the correct imposter. The most effective message that a crewmate could send would convince other crewmates to vote out the true imposter. 

%Note that having accurate, calibrated beliefs on the identity of the imposter is crucial for a strong learning signal, which is a key reason for using the focal loss in \cref{eq:focal} over the more common cross entropy loss. If beliefs are not calibrated, the signal for the effectiveness of a message is noisy and hard to learn from.

Using speaking and listening, we can train an agent $\pi_\text{RL+S+L}$ that minimizes the following loss:

\begin{equation}
    L_\text{RL+L+S}(\pi) = L_\text{RL+L}(\pi) - \underset{\tau^i \sim \Pi}{\mathbb{E}}\sum_t[\lambda_\text{S} \gamma^t r^s_t].
    \label{eq:full_loss}
\end{equation}

\subsection{Training for Dynamic Settings}

As a team zero-sum game, we want our trained crewmates to work well against a wide range of imposters. To do so, we employ an iterated self-play algorithm, where crewmates and imposters train against earlier iterations of their adversary's policy. We train imposters to learn to mislead crewmates into voting out other agents, so we keep the RL loss and invert the speaking loss, minimizing the following:

\begin{equation}
    L_\text{imp}(\pi) = L_\text{RL}(\pi) + \underset{\tau^i \sim \Pi}{\mathbb{E}}\sum_t[\lambda_\text{S} \gamma^t r^s_t].
    \label{eq:imp_loss}
\end{equation}

As the inner optimization loop, we use independent PPO~\citep{schulman2017proximal, huang2022cleanrl} with shared networks for policy and value functions and the Schedule Free AdamW optimizer~\citep{defazio2024schedulefree}.

We also want our crewmates to be robust to different partners who also act reasonably. Therefore, we always set one crewmate to be frozen to the listening policy $\pi_\text{L}$ when forming the joint policy $\Pi$, following the N-Agent Ad hoc teamwork setting~\cite{wang2024naht} instead of assuming a homogeneous population. This change also ensures that crewmates cannot simply determine the identity of the imposter by forming an arbitrary convention and voting out any agent who violates that convention. 

Finally, we want our agents to be robust to different environment configurations. We randomize multiple environment parameters while training: choosing between three different layouts of the environment ($1 \times 3, 2 \times 2$, and $2 \times 3$ grids), and randomizing the number of tasks assigned to each crewmate to either 3, 4, or 5. We only train on configurations where there are 4 crewmates and 1 imposter, but we report generalization results when playing with different numbers of crewmates.%We treat the $2 \times 2$ layout with 4 tasks and 5 players to be the ``base'' environment, but we additionally test deviations from this base environment in our results.

To stabilize training, we also include the following world modeling loss to each model's loss function:

\begin{equation}
    L_\text{WM}(\pi) = -\underset{\tau^i \sim \Pi}{\mathbb{E}}\sum_t \lambda_\text{WM}\log \pi(o^i_{t+1} | \tau^i_t, a^i_t),
    \label{eq:wm}
\end{equation}

where $\lambda_\text{WM}$ is the relative strength of the world modeling loss. Although this loss does not directly contribute to improving task success, it subtly helps improve the model's performance. In particular, as a recurrent model, RWKV benefits from this world modeling loss as it ensures that features are remembered throughout training. Furthermore, the world modeling loss prevents the model from placing too much weight on action tokens, which would cause models to output action tokens even during regular discussion sections.

% We train our models using the Team-PSRO algorithm~\citep{mcaleer2023teampsro}, which learns approximate team-maxmin equilibrium with coordination device (TMECor):

% \begin{equation}
%     \arg \max_{\pi} \min_{\pi_\text{imp}} \mathbb{E}_{r_t^i \sim D(\pi_{\text{imp}}, \pi_{\text{L}}, \pi)}[\gamma^t r_t^i]
% \end{equation}

% This setup assumes that all players can coordinate before playing in the game, but imposters are also aware of the coordination strategies, so the agents cannot determine who the imposter is before the game. 

\section{Results}

\begin{figure}
    \centering
    \includegraphics[width=0.9\columnwidth]{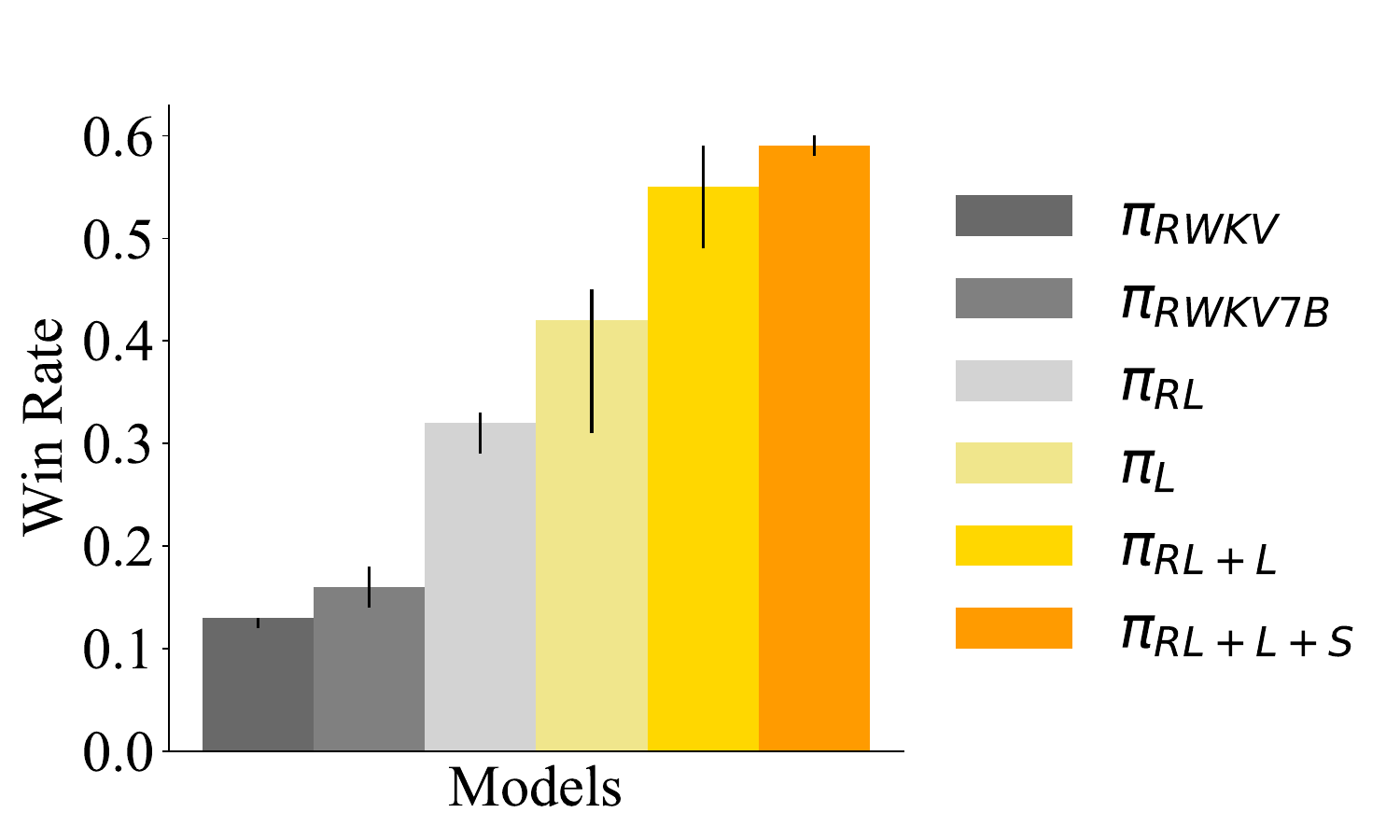}
    \caption{Win rates for crewmates trained with different algorithms over the ``base'' environment: $2 \times 2$ grid of rooms, 4 tasks per crewmate, and 5 players. Error bars represent the maximum and minimum expected win rates across the three independently trained runs with different seeds.}
     % \vspace{-20pt}
    \label{fig:base_env_figures}
    \Description{Win rates for crewmates in the base environment over different algorithms.}
\end{figure}

\begin{figure*}
    \centering
    \includegraphics[width=0.95\textwidth]{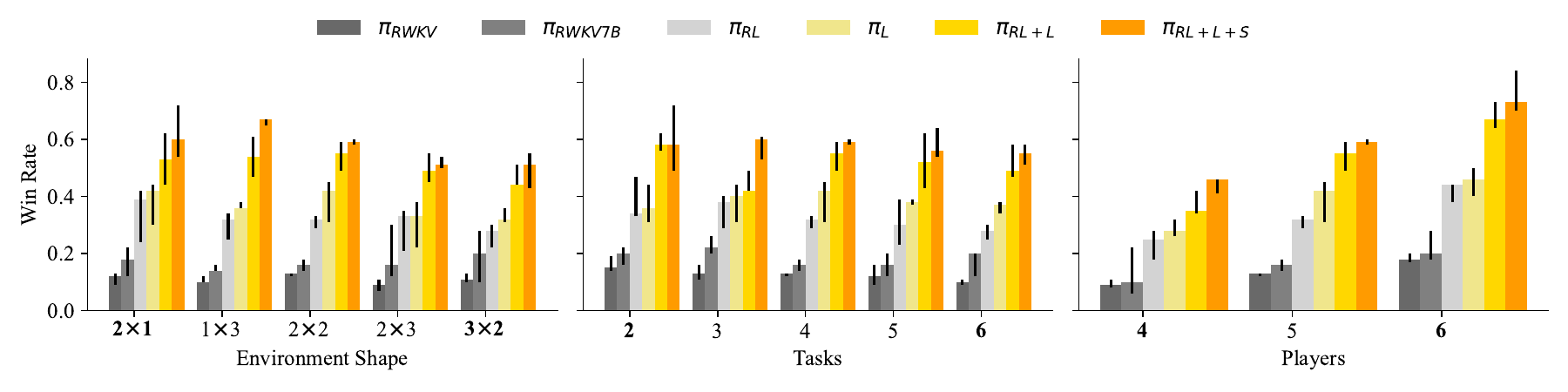}
    \caption{Win rates for crewmates trained with different algorithms over different configurations of the environment, modifying the environment shape, tasks, and number of players.}
    \label{fig:cooperative_figures}
    \Description{Win rates for crewmates across different environment configurations and algorithms.}
\end{figure*}

In this section, we analyze the quality of our trained crewmates. We inspect the importance of different design decisions regarding the training of crewmates by ablating over the components of the loss function in \cref{eq:full_loss}: RL, listening, and speaking. We determine the importance of discussions in the game by measuring the equilibrium win rates of crewmates against imposters and analyze emergent behaviors in discussions. Finally, we highlight some common failure modes we observed during training and how they are mitigated in our final training algorithms.
%We detail results of the base models in~\cref{app:base} and the robustness of agents across different environment parameters in~\cref{app:robust}.

\subsection{Cooperative Training}
\label{sec:coop}

For this set of experiments, we analyze the performance of the crewmates from the first iteration of the self-play algorithm. We conduct all experiments using the 1.5B RWKV model, because we find diminishing returns at higher parameter counts from our experiments on base models (see Appendix B %\cref{app:base}
for more details). We report the win rates of each policy in the base environment when keeping the imposter and one crewmate fixed to $\pi_\text{L}$ in \cref{fig:base_env_figures}.
% For this set of experiments, we keep the imposter and one crewmate fixed to the base model and only train the remaining crewmates, which corresponds to the first iteration of our self-play algorithm. 

\smallskip

\noindent \textbf{Model Evaluations.}
The simplest baselines are direct evaluations of the base model. We find that the 1.5B RWKV model struggles to win in the game, with the larger 7B parameter model performing slightly better as both win less than 20\% of the time in the base environment.

Just training with RL significantly boosts the performance relative to the base models, even significantly outperforming the 7B parameter model. However, we still find that RL without the additional listening loss struggles to reason about the identity of imposters. Even when warm-starting the RL policy from $\pi_\text{L}$, we find that it quickly loses the ability to identify imposters, instead voting for any agent with equal probability. When we instead only trained with listening -- using the loss $L_\text{L only}$ -- the model, $\pi_\text{L}$, does not know which actions are effective or how to discuss details about the environment, but it is an effective baseline due to the fact that predicting the identity of the imposter is valuable in \acro.

When combining RL and the listening loss, we find success rates again increase dramatically, with further improvements when adding our denser \textit{speaking} rewards, as agents can now differentiate between helpful and unhelpful messages when training. We ultimately find that our full model achieves twice the win rate of the RL-only baseline in the base environment. Note that the difference in scores when adding the additional speaking term is relatively small. Even without the explicit speaking reward, the language model produces coherent messages, often sharing their current suspicions during discussion rounds, thus benefiting from discussion even without additional rewards. This is an interesting emergent behavior as it shows that speaking is indirectly improved by training the model to listen better.

% We also find strong out of distribution performance for our models in \cref{fig:cooperative_figures}, indicating that the discussion strategies are robust to changes in the environment; we provide a more detailed analysis in \cref{app:robust}.

\smallskip

\noindent \textbf{Robustness to Environment Variation.}
We present the win rate of crewmates against imposters across different environment configurations in \cref{fig:cooperative_figures}, and see that the trends between models observed in the base environment generally persist across configurations. We find that the shape of the environment has little effect on the win rates of crewmates, with smaller environments generally being easier since it is harder for imposters to kill crewmates without witnesses. We see a general decline in performance across all models when increasing the number of tasks, because this makes it harder to win the game by completing tasks instead of voting out the imposter. Finally, we see a significant increase in win rates as the number of crewmates increase, which we expect since the crewmates can still recover from incorrectly voting out a crewmate.

We do not observe significant deviations from the expected trend lines in settings that were out of the training distribution, demonstrating how the language models can extrapolate their behaviors to unseen deviations in the configuration. 

%We also find that having fewer than four tasks per crewmate results in higher win rates, but there is little difference between 4, 5, and 6 tasks, implying that crewmates typically win by voting out the imposter instead of completing tasks under these configurations. Finally, we find that win rates increase with more players, which we expect since the crewmates can still recover from incorrectly voting out a crewmate.

\smallskip

\noindent \textbf{Message Evaluations.}
We find a major difference between the message patterns of the base RWKV model and those from $\pi_\text{RL+L+S}$. Most messages from the base RWKV model are often unfocused, hallucinating a wider context to the game and role-playing a crewmate. Meanwhile, crewmates using $\pi_\text{RL+L+S}$ often directly accuse the imposter or otherwise name the imposter in their messages. In general, we find that naming an agent makes it more likely for other agents to vote against them. Furthermore, crewmates share messages that resemble environment observations that helped them judge the identity of the imposter. For instance, a crewmate may say ``Player Green is leaving Room (0,1)'' when the body is in Room (0,1) to indicate that Player Green was running away from the dead body, which is often correlated with being the imposter. However, the crewmates sometimes tell lies in their messages -- just like humans often do when playing \acro. In particular, they often simply make up evidence and state whatever is most convincing to other agents to get enough votes to eject the correct imposter. Representative behavior samples are provided in Appendix C.%\cref{app:sample_generations}.

\begin{figure}
% \vspace{-10pt}
  \centering
  \includegraphics[width=0.9\columnwidth]{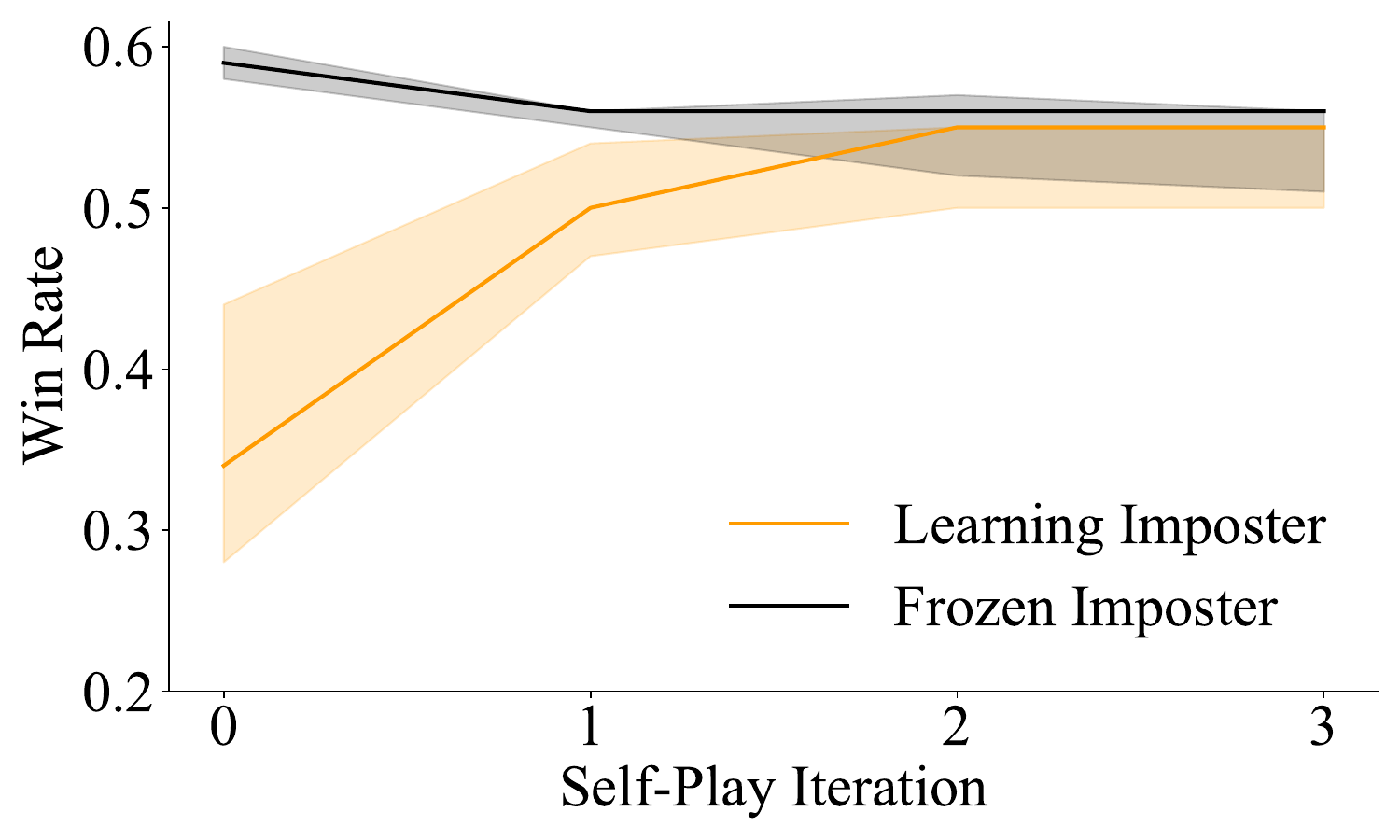}
  \caption{Exploitability curves for policies over self-play iterations, evaluated on the base environment. The orange line indicates the expected win rate against an adversarially trained imposter. The black line indicates the expected win rate of crewmates who are specifically optimized against this iteration's imposters. Note that iteration 0 refers to the base models, while iteration 1 refers to the crewmate policy from the Cooperative Training section. Shaded regions represent the maximum and minimum win rates across the three independently trained runs with different seeds.}
    \label{fig:psro}
    \Description{Graph of crewmate win rate over self-play iterations}
\end{figure}

\subsection{Robustness to Imposters}
When training against frozen imposters, crewmates could come up with simple strategies that result in high win rates but are easy to overcome with a more intelligent imposter. We therefore run multiple iterations of self-play to investigate whether crewmates can use discussions even against imposters that can evolve to their policies, which we illustrate in \cref{fig:psro}. Note that in exploitability curves, we would like to see convergence upon a narrow interval for both the upper and lower bounds; a weak strategy can be easily exploited at each iteration and would therefore have both lines stay far apart and converge slowly.

We find that the crewmates' strategies are robust to imposters trained in an adversarial fashion. In particular, we see that crewmate scores converge after only a few iterations; depending on the seed, the win rate converges to between 0.51 and 0.56 on the base environment. In fact, even the crewmates that only trained on the base model imposter are relatively strong, as can be seen by the large jump in the lower bound between iterations 0 and 1. This result implies that policies discovered by $\pi_\text{RL+L+S}$ are very robust since even facing adversarially trained imposters does not cause a significant performance drop. 

Qualitatively, we observe that imposters attempt to shift blame to other players by counter-accusing another crewmate. In particular, they mimic the discussion patterns of crewmates, and the crewmates sometimes fall for this deception. Crewmates who have not witnessed the murder tend to support claims made by other players, causing them to sometimes help the imposter. Interestingly, we still see similar behavior to a smaller level in the base model imposters when playing against strong crewmates. This emergent behavior can likely be attributed to the in-context learning capabilities of language models, which would allow the imposter to mimic the speech of crewmates who spoke beforehand.

\subsection{Failure Modes}

Throughout our experimentation, we encountered various failure modes that we tackled in our final training algorithms. Specifically, discussions tended to leave natural language and generally degenerate without careful consideration from the training algorithm.

First, we observed that the soft KL constraint commonly used in RLHF~\cite{ouyang2022training} required careful tuning to keep language generations in English. When this constraint is weighted too low, all of our RL-trained models diverge from natural language after only a few iterations, causing it to output random tokens during discussions and stop improving in performance.

We also observed that allowing all crewmates to be trained simultaneously would lead to degenerate solutions. Sometimes the models learn a social convention where they simply do not speak during the discussion phase. Specifically, models would output newlines when it is their turn to speak instead of actually speaking. In this case, only the imposter would speak and all the agents would just vote the speaker out. The models would also learn to just wait in the starting room instead of moving around, allowing them to witness the murder or vote out the person who moves out of the room. These strategies are degenerate solutions since they would not work if the imposter was aware of their strategy or if not all the crewmates shared the same strategy, but these strategies would lead to nearly perfect win rates during the first iteration of self-play. The fix to this issue was to ``freeze'' one crewmate to not learn and therefore not follow changes in strategies. 

The final failure mode we observed was using action tokens in discussions instead of natural language. Specifically, the RL-trained models would learn to take actions by explicitly choosing action tokens, but this gives action tokens a higher probability to be chosen overall, even during discussion phases. We observed that the best way to counteract this effect was to introduce the world modeling loss from~\cref{eq:wm}. This loss ensured that the model preserved its language modeling abilities and had the side effect of helping the models match the patterns it experienced in observations within its own discussions, which would help independent agents understand the intentions of our models.

\section{Discussion}
\noindent \textbf{Summary.}
We introduce a technique to self-improve the discussion ability of an LLM in social deduction games and show how it enables agents to communicate effectively in the game of \acro. We demonstrate that, despite having weak base models, our agents learn to speak effectively and extract information from discussion messages. We also find that our agents are robust to adversarially trained imposters, who, despite attempting to sabotage the discussion, are unable to break the crewmates' coordination during discussions. Our technique ultimately shows that self-improving discussions in multi-agent settings does not require task-specific human data, unlocking the possibility for multi-agent communication with language models in novel tasks.

\smallskip

\noindent \textbf{Limitations and Future Work.}
% Our LLMs were based on the RWKV-4 set of pretrained models due to its ability to be expressed as both a recurrent network and a linear transformer, enabling us to efficiently generate RL buffers and train our models over very long context lengths. However, other architectures may behave differently in our environment and follow different scaling laws.
A key limitation of our approach is that our scene prediction technique is task-dependent. In \acro, there is a natural connection between the discussion and trying to predict the identity of the imposter, and a similar structure applies to a wide range of social deduction games and real-world settings. An interesting future direction would be to allow agents to identify which aspects of the scene are relevant to a specific task instead of manually specifying it. Please refer to Appendix D for more analysis on the broader impacts of our work.

We also note that crewmates are not always truthful in their discussions, opting instead to make the most convincing statements. We consider this behavior to be potentially dangerous outside of our sandboxed setting of Among Us, so we believe that optimizing for truthfulness is an important future direction.

%%%%%%%%%%%%%%%%%%%%%%%%%%%%%%%%%%%%%%%%%%%%%%%%%%%%%%%%%%%%%%%%%%%%%%%%

%%% The acknowledgments section is defined using the "acks" environment
%%% (rather than an unnumbered section). The use of this environment 
%%% ensures the proper identification of the section in the article 
%%% metadata as well as the consistent spelling of the heading.

\begin{acks}
This research was supported in part by the Other Transaction award HR00112490375 from the U.S. Defense Advanced Research Projects Agency (DARPA) Friction for Accountability in Conversational Transactions (FACT) program, the Cooperative AI foundation, ONR project N00014-21-1-2298, NSF Awards \#2006388, \#1941722, \#2125511, AFOSR YIP, and the Stanford Center for Human-Centered AI (HAI). We thank the Stanford Madrona team for giving advice on the systems-level details of our implementation, and Hengyuan Hu for his insightful feedback when reviewing this paper.
\end{acks}

%%%%%%%%%%%%%%%%%%%%%%%%%%%%%%%%%%%%%%%%%%%%%%%%%%%%%%%%%%%%%%%%%%%%%%%%

%%% The next two lines define, first, the bibliography style to be 
%%% applied, and, second, the bibliography file to be used.

\bibliographystyle{ACM-Reference-Format} 
\bibliography{main}

%%%%%%%%%%%%%%%%%%%%%%%%%%%%%%%%%%%%%%%%%%%%%%%%%%%%%%%%%%%%%%%%
%% Appendices
%%%%%%%%%%%%%%%%%%%%%%%%%%%%%%%%%%%%%%%%%%%%%%%%%%%%%%%%%%%%%%%%
\newpage
\clearpage
\appendix

\onecolumn

\section{Environment Design}
\label{app:env}

\smallskip

\noindent \textbf{Gameplay Phase.} The main gameplay loop consists of players navigating a 2D environment, consisting of a $W \times H$ grid of rooms. The location of the agent is just the room number; movement within the room takes no time. All agents can move between rooms based on a set speed of movement, $N_{travel}$. 

Crewmates use this time to complete ``tasks'' around the map. In the original game of \acro, tasks involve completing minigames that require a player's attention, such as solving a maze, preventing them from observing the environment around them. In our implementation, we simplify the notion of tasks, and require the crewmates complete tasks by only staying in the room containing the task for a specified amount of time. However, similar to the original game, crewmates receive no observations from the environment while completing the task, which leaves them vulnerable to attacks and may cause them to miss out on information such as an imposter killing another crewmate. If all tasks are completed, the crewmates win and the game ends, which incentivizes them to attempt performing tasks despite the risks and partial observability that comes with it.
% This design choice allows our version of the game to keep the core reason for having tasks.

Imposters are not assigned tasks to complete, and instead use the main gameplay loop to eliminate crewmates. Eliminating a crewmate is an action where the imposter kills a specific crewmate in their current room. Imposters have to wait for an ``elimination cooldown'' before eliminating a crewmate, and this countdown restarts after each elimination, preventing imposters from instantly eliminating all crewmates. When a crewmate is eliminated, their corpse is left behind, and any player who finds the corpse can report it and instantly start the discussion phase.
% , described in the next subsection.

Since the environment interfaces with agents through natural language, the environment constructs observation descriptions based on the surroundings of the agent. Each observation $o^i_t$ include the current timestamp of the environment, the current room that the agent is in, and the list of other players in the room. Note that this observation does not include if other players are completing a task as waiting in a room looks identical to working on a task. If another player is travelling towards or away from the current room, this is also indicated. Here is an example of an observation for a crewmate at timestep 56: 
\begin{verbatim}
[56]: You are in room (0, 1). You see Player Green leaving to room 2. You have the following tasks in this room: Task 2.
\end{verbatim}

Furthermore, crewmates are given information about uncompleted tasks in the current room, and imposters are given the number of seconds remaining in the elimination cooldown.

Following an observation, the agent also receives a discrete set of legal actions based on the state, which they must pick from. Here is an example of the action set:

\begin{verbatim}
[56] World: You can perform any of the following actions: go north; wait; do task; go south; wait
[56] You: wait
\end{verbatim}

%The list of all actions an agent can take is provided in Table~\ref{tab:validactions}. 
All agents are allowed to just ``wait'' in a room until something changes in the environment, or ``go'' to an adjacent room, taking time to travel. If there is a corpse near an agent, they can ``report body'' and initiate the discussion phase. Crewmates can ``do task'' to do an uncompleted task in the room, while imposters can ``kill player [x]'' to eliminate a specific player if they are not on the elimination cooldown. Note that although imposters may not complete tasks, they can still appear to do tasks to an outside observer by simply performing the ``wait'' action. To enable efficient action generation with language models, each action is mapped to a unique token in a multiple-choice format.%, represented by the letter preceeding each action in Table~\ref{tab:validactions}.

% \begin{table}[htbp]
%     \begin{center}
%         \begin{tabular}{l|l|lll}
%             \multicolumn{1}{c}{\bf Imposter Only}  &\multicolumn{1}{c}{\bf Crewmate Only} &\multicolumn{3}{c}{\bf Any Agent}\\ 
%             \hline \\
%             B kill Player 0 & L do Task 0 & G report body of Player 0 & Q enter room 0 & W wait \\
%             C kill Player 1 & M do Task 1 & H report body of Player 1 & R enter room 1 & \\
%             D kill Player 2 & N do Task 2 & I report body of Player 2 & S enter room 2 & \\
%             E kill Player 3 & O do Task 3 & J report body of Player 3 & T enter room 3 & \\
%             F kill Player 4 & P do Task 4 & K report body of Player 4 & U enter room 4 & \\
%             &&& V enter room 5&
%         \end{tabular}
%     \end{center}
%     \caption{List of all 22 actions during the \acro gameplay phase.}
%     \label{tab:validactions}
% \end{table}

\smallskip

\noindent \textbf{Discussion Phase.} When an agent reports the body of another player, the discussion phase starts and all players are immediately brought to a central room. The environment informs all players about the identity of the corpse and the player who reported the body. 

To determine the order of speakers, the environment cycles through a randomized list of living players twice. During a player's turn to speak, they can produce up to 20 tokens or until a newline character, and this message is shared with all agents before the environment chooses the next player. Unlike the gameplay phase, where actions are restricted to a small set of tokens, discussions are free-form so we allow agents to generate any printable token. We allow up to 20 tokens since this roughly corresponds to the maximum number of tokens used in the longest messages in the ``quick chat'' setting of the original \acro game. In practice, trained agents tend to use fewer than 20 tokens per message, instead ending their turn early using newline characters.

 % \karen{Did we have any ablations on the parameters of the game design? For example, how do all these arbitrarily determined numbers like 6 messages or 20 tokens impact our results?}.

After the free-form discussion, a voting phase begins. Each player is given the option to vote to eject a living player or abstain. The agent who achieves the plurality of the votes gets ejected, except in the case of ties or when ``abstaining'' wins, at which point nobody gets ejected. If all imposters are ejected, the crewmates win and the game ends. Otherwise, the gameplay phase starts again and the cycle continues.

Before the discussion begins and between each discussion message, the environment queries all living crewmates to ask who they believe is the current imposter, and collects their probabilities based on the language model's probabilities of voting out each agent. This has no impact on the game itself, but it is used for training.

\smallskip

\noindent \textbf{Reward Structure.} \acro is fundamentally a zero-sum team game, so the reward is based on whether crewmates or imposters win. If all tasks are completed or all imposters are ejected, the crewmates win, representing a crewmate reward of $+1$. If the number of imposters is ever greater than or equal to the number of crewmates, the imposters win, resulting in a crewmate reward of $-1$.
% \karen{At this point, as a reader who has not played Among Us before, I am concerned about the sensitivity of the parameters in the game mechanism and how arbitrarily we determine them. It seems that one of the most important game parameters is the initial ratio of the imposter to crewmates. If the ratio is high, the game is significantly more difficult for crewmates because the imposters can lie for each other and can easily out vote uninformed crewmates.}

% \karen{Also, at this point, I still don't know what our success metric is. Does success mean that crewmates have a higher win rate or imposters have a higher win rate? Or it means that we can show that the agents reach an optimal equilibrium?}

\pagebreak

\section{Base Model Results}
\label{app:base}

The original RWKV models came in sizes ranging from 169 million parameters to 7 billion parameters. To determine the scaling laws of these models on our environment, we trained our base models using \cref{eq:base} over the same amounts of randomized environment interaction data. We present the results of the trained models in \cref{tab:baseModelAblation}. For the metric of accuracy, we find diminishing returns at 1.5 billion parameters, so we use this model for ablation experiments.

% \begin{figure}
%     \centering
%     \begin{minipage}[b]{.48\textwidth}
%          \centering
%   \includegraphics[width=\textwidth]{figures/BaseTrainingCurves.png}
%   \caption{Accuracy of supervised imposter prediction for the base model over training time for different parameter sizes.(Note: need to update)}
%     \label{fig:basetraining}
%     \end{minipage}%
%     \hfill
%     \begin{minipage}[b]{.48\textwidth}
%         \centering
%         \includegraphics[width=0.5\textwidth]{figures/CrewmateOnly.png}
%   \caption{Win rates of crewmate-only training. The orange curve indicates win rates when discussions are allowed while the gray curve does not allow discussions. (Note: placeholder)}
%     \label{fig:crewmateonly}
%     \end{minipage}
% \end{figure}

\begin{table}[htbp]
    \begin{center}
        \begin{tabular}{l|lll|lll}
        \multicolumn{1}{l}{} & \multicolumn{3}{c}{\bf Our Base Model} & \multicolumn{3}{c}{\bf Original RWKV} \\
            \multicolumn{1}{l}{\bf Model Size}  &\multicolumn{1}{l}{\bf Accuracy $\uparrow$} &\multicolumn{1}{l}{\bf PPL$_{\text{World}}$ $\downarrow$}&\multicolumn{1}{l}{\bf PPL$_{\text{Lam}}$ $\downarrow$} &\multicolumn{1}{l}{\bf Accuracy $\uparrow$} &\multicolumn{1}{l}{\bf PPL$_{World}$ $\downarrow$}&\multicolumn{1}{l}{\bf PPL$_{Lam}$ $\downarrow$}\\ 
            \hline \\
            169 M & 0.60 & 0.34 & 45.1 & 0.15 & 4.4 & 27.5 \\
            430 M & 0.61 & 0.30 & 19.6 & 0.18 & 2.9 & 11.2 \\
            1.5 B & 0.62 & 0.28 & 7.3 & 0.20 & 2.4 & 6.2 \\
            3 B   & 0.64 & 0.23 & 6.2 & 0.14 & 2.0 & 4.8 \\
            7 B   & 0.65 & 0.22 & 5.2 & 0.22 & 2.0 & 4.1 \\
        \end{tabular}
    \end{center}
    \caption{Scaling performance for the RWKV base model. PPL$_{\text{World}}$ refers to the perplexity of predicting observation tokens in the environment, which correlates with a model's ability to understand its environment. PPL$_{\text{LAM}}$ refers to the lambada evaluation perplexity.}
    \label{tab:baseModelAblation}
\end{table}

% \section{Robustness to Environment Configuration}
% \label{app:robust}

% When training, we randomize the shape of the grid that agents navigate during gameplay and the number of tasks that each crewmates is assigned. We present the win rate of crewmates against imposters across these different configurations in \cref{fig:cooperative_figures}, and see that the trends between models observed in the base environment generally persist across configurations. We find that the shape of the environment has little effect on the win rates of crewmates, with smaller environments being marginally easier since it is harder for imposters to kill crewmates without witnesses. We also find that having fewer than four tasks per crewmate results in higher win rates, but there is little difference between 4, 5, and 6 tasks, implying that crewmates typically win by voting out the imposter instead of completing tasks under these configurations. Finally, we find that win rates increase with more players, which we expect since the crewmates can still recover from incorrectly voting out a crewmate.

\section{Sample Generations}
\label{app:sample_generations}

All samples below are from the perspective of the imposter:

\medskip
Sample 1: Through the discussion, Player Pink learns that Player Blue is the imposter:
\begin{verbatim}
World (to all): Player Yellow discovered the dead body of Player Purple in room (0,0).
Player Blue (you) saying: "There's no way I'm playing this game."
Player Yellow (to all): "Only the Imposter will speak."
Player Green (to all): "The other Crewmates are suspicious of Player Blue. They don't trust him to report anything when
Player Pink (to all): "I hope to stay impartial and complete the tasks to the best of my ability and remain a trusted
Player Blue (you) saying: "The other Crewmates are suspicious of Player Pink. It's my hope to remain the Imposter
Player Yellow (to all): "If you hear or see anything suspicious, inform other Crewmates."
Player Green (to all): "I notice suspicious activities of Player Blue and Player Purple."
Player Pink (to all): "I notice suspicious activities of Player Blue."
Player Blue was voted out. Crewmates win!
\end{verbatim}

\medskip
Sample 2: Other players directly accuse Player Green:
\begin{verbatim}
World (to all): Player Purple discovered the dead body of Player Yellow in room (2,0).
Player Orange (to all): "The Imposter is Player Green."
Player Green (you) saying: "Shut Up!"
Player Purple (to all): "The Imposter is Player Green."
Player Orange (to all): "You are the Imposter!"
Player Green (you) saying: "My strategy is to kill or eject Imposters."
Player Purple (to all): "Players can only report each other and identify the Imposter."
Player Green was voted out. Crewmates win!
\end{verbatim}

\medskip
Sample 3: Player Pink, the imposter, convinces the other crewmate to vote out the innocent Player Orange.
\begin{verbatim}
World (to all): Player Purple discovered the dead body of Player Red in room (1,0).
Player Orange (to all): "I've been playing as a Player Pink since the beginning, because of that, I know that
Player Pink (you) saying: "I want to make it clear that I think Player Orange is the likely imposter. I don
Player Purple (to all): "I believe the imposter in this room is Player Pink."
Player Orange (to all): "I believe that Player Purple is the imposter in this
Player Pink (you) saying: "I believe the imposter is Player Cubie."
Player Purple (to all): "I believe Player Pink killed Player Orange."
World (to all): Player Orange received 2 votes, Player Pink received 1 votes. Therefore, Player Orange is ejected 
                this round.
World (to all): There are currently more imposters than crewmates. Imposters win!
\end{verbatim}

\medskip

Sample 4: Player Green says false information, but the imposter is still voted out.
\begin{verbatim}
World (to all): Player Yellow discovered the dead body of Player Purple in room (2,0).
Player Green (to all): "I don't know."
Player Blue (to all): "I am not sure," "I do not know," "I do not recognize the dead body
Player Red (you) saying: "I believe Player Green is the Imposter."
Player Yellow (to all): "It is our obligation to investigate whether the dead body of Player Red is the Imposter."
Player Green (to all): "The dead body in room (1,0). Player Red killed Player Blue in room (1
Player Blue (to all): "I believe Player Red is the Imposter."
Player Red (you) saying: "The dead body in room (1,0) is the Imposter. There are currently more
Player Yellow (to all): "I believe Player Red is the Imposter."
Player Red was voted out. Crewmates win!
\end{verbatim}

\medskip

Sample 5: Despite counter-accusing, Player Orange is voted out by the rest of the group.
\begin{verbatim}
World (to all): Player Yellow discovered the dead body of Player Blue in room (1,1).
Player Red (to all): "It is clear Player Orange is the Imposter. Therefore, I vote for Player Orange to be
Player Yellow (to all): "I'm Player Orange, Player Red knew Player Orange. Player Orange killed Player Red."
Player Orange (you) saying: "The Player Red is the most suspicious Player"
Player Red (to all): "I am Player Orange"
Player Yellow (to all): "I vote Player Orange"
Player Orange (you) saying: "I vote Player Yellow"
\end{verbatim}

\medskip

Sample 6: Agents say evidence they observed in the environment.
\begin{verbatim}
World (to all): Player Pink discovered the dead body of Player Green in room (1,0).
Player Pink (to all): "I have a suspicion that Player Red killed Player Purple. There are currently more Crewmates 
                      than 1
Player Purple (you) saying: "I think it is the Player Red in the room."
Player Red (to all): "From Player Purple: The Crewmates discovered the dead body of Player Purple in room (1,
Player Pink (to all): "I think Player Red is the Imposter."
Player Purple (you) saying: "I think it is Player Red.
Player Red (to all): "I think I see Player Purple leaving from room (0,0).
World (to all): Player Red received 1 votes, Player Purple received 1 votes, Player Pink received 1 votes. 
                Therefore, nobody is ejected this round.
\end{verbatim}

\section{Broader Impacts}
\label{app:impact}

The primary goal of this work is to enable multi-agent communication between language model agents in novel settings without human demonstration data. Although the discussions generated by our agents are still relatively simple, our technique has the potential to scale to larger models with sufficient compute.

Strong multi-agent communication could have great benefits, such as deploying cooperative agents in settings where humans cannot reasonably provide demonstrations, like at microscopic scales or in toxic environments. Being able to specifically communicate in natural language would also be very useful, since it can help enable smoother human-AI coordination in heterogeneous teams.

There are also potential risks to deploying our technique outside of \acro. In particular, we notice that both crewmates and imposters make statements that are not backed by evidence. It is unclear whether this is simply a result of using small models that are lacking in the ability to recall information precisely or if this is a fundamental feature that will be preserved regardless of scale. We encourage future researchers to continue studying \acro and similar sandboxed settings to understand the multi-agent dynamics of these language models before deploying large-scale multi-agent learning systems that can interface with the real world.

\section{Hyperparameters and Compute}
\label{app:compute}

We use the AdamWScheduleFree optimizer from~\citet{defazio2024schedulefree} so we don't have a separate scheduler.

\begin{table}[H]
\centering
\caption{Common hyperparameters}
\vskip 0.15in
\begin{tabular}{cc}
\toprule
hyperparameters & value \\
% Species & CBS & CV & G3 \\
\midrule
lr & 3e-4 \\
$\lambda_\text{BC}$ &  1.0\\
$\lambda_\text{WM}$ &  1.0\\
$\lambda_\text{NL}$ &  0.05\\
$\lambda_\text{L}$ &  0.3\\
$\lambda_\text{S}$ &  1.0\\
\bottomrule
\end{tabular}
\vskip -0.1in
\end{table}

An exception to the above hyperparameters is that $\lambda_\text{L}=3.0$ for $\pi_\text{RL+L+S}$ and $\lambda_\text{L}=0.1$ for $\pi_\text{RL+L}$ because we find that it significantly impacts stability. We use a batch size of 30 environments when collecting RL trajectories, but we subdivide them into processing 6 trajectories in parallel during optimization.

All experiments were conducted on individual A40 GPUs with 48 GB of VRAM. All models can be trained within 48 hours of compute.

\section{Assets and Licenses}
\label{app:license}

We borrow code from CleanRL's PPO implementation~\cite{huang2022cleanrl}, provided under the MIT license. 

We draw inspiration from Innersloth's Among Us game, which gives permission to use the Among Us IP for non-commercial and educational use. Our work is not associated with or officially licensed by Innersloth. Depictions of Among Us characters in \cref{fig:amongus_diagram} are for illustrative purposes.

All art assets in this paper were created using Processing, Matplotlib, and Keynote. 

This paper is provided to the public under the CC-BY-4.0 License, and all associated code is shared under GNU GPL v3.0.

%%%%%%%%%%%%%%%%%%%%%%%%%%%%%%%%%%%%%%%%%%%%%%%%%%%%%%%%%%%%%%%%%%%%%%%%

\end{document}